\documentclass[conference]{IEEEtran}

 \usepackage[dvipsnames,table,xcdraw]{xcolor}
\usepackage{pgfplots}
\usepackage{pgfplotstable}
\usepackage{booktabs}
\usepackage{colortbl}
\usepackage{float}
\usepackage{url}
\usepackage{booktabs}
\usepackage{pdfpages}
\usepackage{svg}
\usepackage{hyperref}
\usepackage{subcaption}
\usepackage{multirow}
\usepackage[utf8]{inputenc}
\usepackage{authblk}
\usepackage[T1]{fontenc}
\usepackage{amsmath,amssymb}
\usepackage{paralist}
\usepackage{latexsym}
\usepackage{wasysym}
\usepackage{graphicx}
\usepackage{MnSymbol}
\usepackage{stackengine}
\usepackage{mdframed}
\usepackage{comment}
\usepackage[ruled,linesnumbered,commentsnumbered,vlined]{algorithm2e}
\usepackage{enumitem}
\usepackage{multicol}
\usepackage{menukeys}
\def\baselinestretch{0.98}

\usepackage{amsmath}
\usepackage{tikz-qtree}

\usepackage[listings,skins,breakable]{tcolorbox}

\usetikzlibrary{shapes} 
\usetikzlibrary{fit,backgrounds}

\SetAlFnt{\scriptsize}

\DeclareMathAlphabet{\mathcal}{OMS}{cmsy}{m}{n}

\renewcommand{\figurename}{Fig.}
\renewcommand{\tablename}{Tab.}

\newcommand{\noleftdelimiter}{\left.\kern-\nulldelimiterspace}

\newcommand*\circled[1]{\tikz[baseline=(char.base)]{
            \node[shape=circle,fill,inner sep=0.5pt] (char) {\textcolor{white}{#1}};}}

\makeatletter
\def\ps@IEEEtitlepagestyle{%
  \def\@oddfoot{\mycopyrightnotice}%
  \def\@evenfoot{}%
}
\def\mycopyrightnotice{%
  {\begin{minipage}{\textwidth}
  \footnotesize \copyright 2021 IEEE. Personal use of this material is permitted. Permission from IEEE must be obtained for all other uses, in any current or future media, including reprinting\slash republishing this material for advertising or promotional purposes, creating new collective works, for resale or redistribution to servers or lists, or reuse of any copyrighted component of this work in other works.
  \end{minipage}
  }
  \gdef\mycopyrightnotice{}
}

\begin{document}

\title{\textbf{CATE}: \textbf{CA}usality \textbf{T}ree \textbf{E}xtractor from Natural Language Requirements}

\author[1]{Noah Jadallah}
\author[2]{Jannik Fischbach}
\author[3]{Julian Frattini}
\author[4]{Andreas Vogelsang}

\affil[1]{\footnotesize Technical University of Munich, Germany, 
noah.jadallah@tum.de}
\affil[2]{\footnotesize Qualicen GmbH, Germany, jannik.fischbach@qualicen.de}
\affil[3]{\footnotesize Blekinge Institute of Technology, Sweden, julian.frattini@bth.se}
\affil[4]{\footnotesize University of Cologne, Germany, vogelsang@cs.uni-koeln.de}

\maketitle

\begin{abstract}
Causal relations (If A, then B) are prevalent in requirements artifacts. Automatically extracting causal relations from requirements holds great potential for various RE activities (e.g., automatic derivation of suitable test cases). However, we lack an approach capable of extracting causal relations from natural language with reasonable performance. In this paper, we present our tool \emph{CATE} (CAusality Tree Extractor), which is  able to parse the composition of a causal relation as a tree structure. \emph{CATE} does not only provide an overview of causes and effects in a sentence, but also reveals their semantic coherence by translating the causal relation into a binary tree. We encourage fellow researchers and practitioners to use \emph{CATE} at \url{https://causalitytreeextractor.com/}
\end{abstract}

\section{Introduction}
\textbf{Motivation:} Black-box behavior of a system is often described by causal relations: \textit{If A and B, then the system shall <functionality>}. Recent studies proved that causal relations occur in traditional requirements documents~\cite{fischbachREFSQ} as well as agile requirement artifacts~\cite{fischbachICST} such as user stories and acceptance criteria. Automatically extracting  causal relations from requirements can help to increase the automation of RE activities. For example, we see great potential for the automatic test case derivation from requirements and the automatic detection of dependencies between requirements~\cite{fischbachRENEXT}. 

\textbf{Problem:} We lack an approach capable of extracting causal relations from natural language with reasonable performance. Existing approaches~\cite{yang2021survey} do not consider the combinatorics (e.g., disjunctions) between causes and effects. They also do not allow splitting causes and effects into more granular text fragments (e.g., variable and condition), making the extracted relations unsuitable for the mentioned use cases.

\textbf{Contributions:} We present our tool \emph{CATE} (acronym for CAusality Tree Extractor), which is able to parse the composition of a causal relation as a tree structure. \emph{CATE} does not only provide an overview of the causes and effects in a sentence, but also reveals their semantic coherence through the binary tree structure. \emph{CATE} is based on Recursive Neural Networks (RNN)~\cite{socher13}, which we trained on the \emph{Causality Treebank} - our self-annotated gold standard corpus of fully labeled binary parse trees representing the composition of 1,571 causal requirements. In this paper, we only present \emph{CATE} as a tool. A detailed description of the \emph{Causality Treebank} and an explanation of why and how we use RNN for causality extraction is available in our AIRE research paper~\cite{fischbachAIRE}.\footnote{Please note that \emph{CATE} is currently only based on the vanilla RNN. We are working on also integrating Recursive Neural Tensor Networks described in our research paper into \emph{CATE}.}

\section{CATE: CAusality Tree Extractor}

\begin{figure}
  \centering
  \scalebox{0.2}{
  \includegraphics{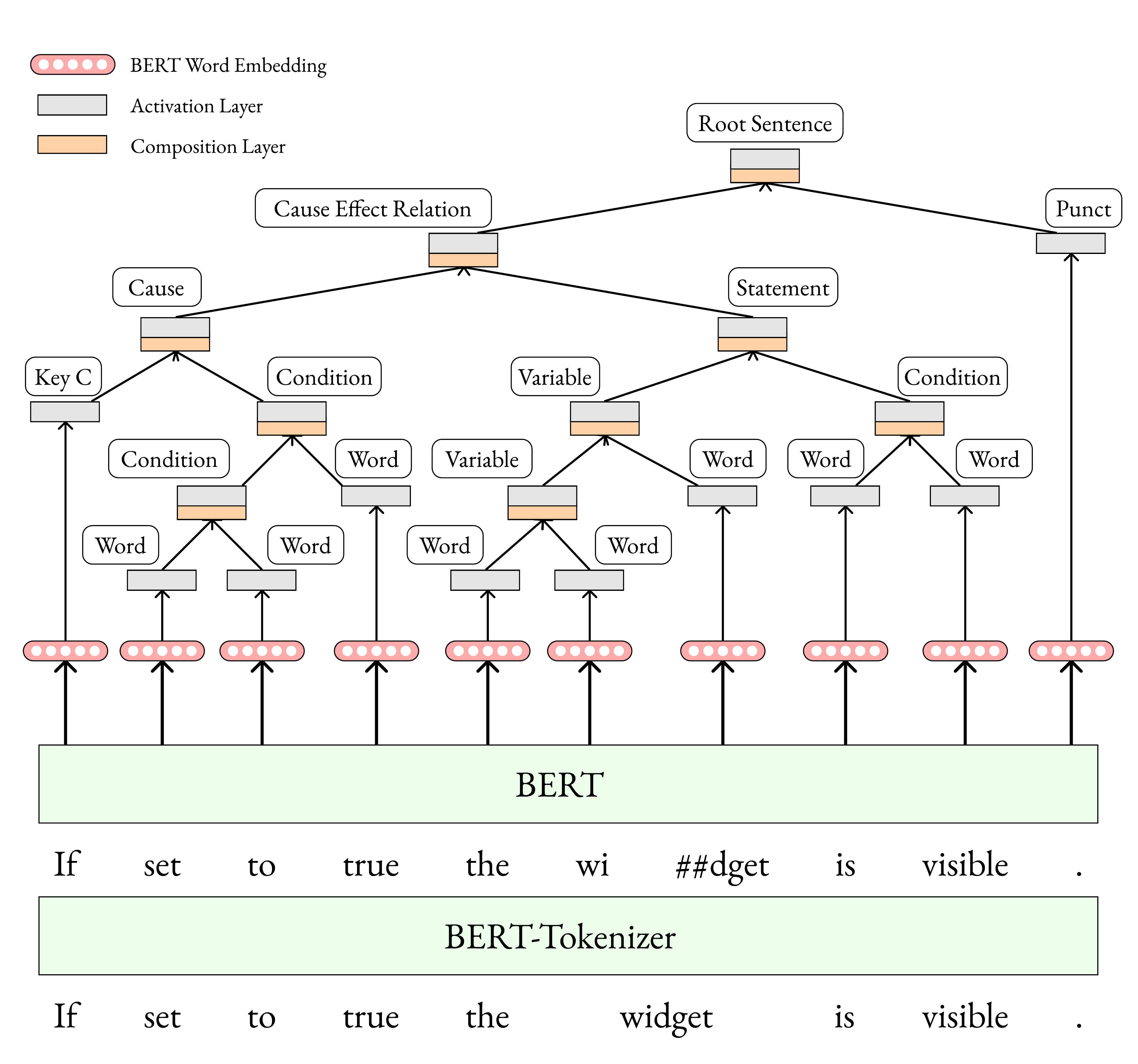}
  }
  \caption{RNN combined with BERT embeddings.}
  \label{bert_architecture}
  \vspace{-0.5cm}
\end{figure}

\textbf{Principal Idea}
A RNN embraces the idea of natural language as a recursive structure. For example, the syntax of a sentence is recursively structured, with noun phrases containing relative phases, which in turn contain further noun phrases, and so on~\cite{socher11}. We understand a causal relation also as a recursive structure, since it consists of causes and effects, which in case of conjunctions and disjunctions consist of further causes and effects, and so on~\cite{fischbachRENEXT}. A RNN can be trained to rebuild this recursive structure. Specifically, it recovers the composition of a sentence by identifying related words and merging them into pre-defined segments (e.g., causes and conditions). This results in a binary tree structure (see Fig.~\ref{bert_architecture}).

\textbf{Left vs. Right-branching}
A RNN builds up the binary tree of segments in a bottom-up fashion and supports two different branching methods: left-branching and right-branching. Let us consider the three tokens ``set to true'', which constitute a condition segment (see Fig.~\ref{bert_architecture}). A RNN can only join adjacent token pairs in each recursion, resulting in two merge options for these tokens. Either ``set'' and ``to'' are merged first (left-branching) and then connected with ``true'' or ``to'' is first connected with ``true`'' (right-branching) and then joined with ``set''. Our first experiments revealed that the RNN seems to learn left-branching better than right-branching~\cite{fischbachAIRE}.

\textbf{Word Embeddings}
To process causal sentences with a RNN, their words must be first converted into a word embedding. \emph{CATE} is trained on two different types of word embeddings: pre-trained word vectors and randomly initialized word vectors that are trained jointly with the other parameters of the RNN. In the case of pre-trained word vectors, we utilize embeddings established in practice and academia: FastText~\cite{fasttext} (1M word vectors), GloVe~\cite{pennington2014glove} (2.2M word vectors) and 768-dimensional BERT embeddings~\cite{bert}.

\textbf{Training vs. Inference}
Emphasis is often put on the training of a model and less on its inference. Since we intend to use \emph{CATE} in practice, we considered two ways of optimizing its inference: \circled{1} temperature scaling and \circled{2} beam search. We compare both methods to the naive approach of creating the binary tree, which can be understood as a simple greedy algorithm. Let us assume a sentence consisting of the tokens: $(a, b, c, d, e)$. The naive approach to create the binary tree for this sentence consists of the following steps:
\begin{enumerate}
\item Compute the softmax scores for all adjacent nodes $\{(a, b), (b, c), (c, d), (d, e)\}$
\item Concatenate the two nodes that achieved the highest posterior probability and update the list of adjacent nodes. 
\item Repeat until only one parent node is left. 
\end{enumerate}

The naive approach runs in linear run-time. However, it holds some major drawbacks which can result in poor parsing results, especially for complex sentences. Firstly, the softmax score can not be interpreted reliably as a confidence score since the predicted probabilities tend to be too high even if the input data does not make any sense. Consequently, the probability associated with the predicted class label does not reflect its ground truth correctness likelihood~\cite{Guo17}. 
As a solution, Guo et al.~\cite{Guo17} propose to use temperature scaling in order to calibrate the predicted probabilities. Specifically, the neural network is complemented with a parameter $T$ in the softmax function, with the objective of achieving confidence scores that closer represent the ``certainty/uncertainty'' of the model. Thus, for a classification into 27 classes, we compute the calibrated softmax score as follows:
\begin{align}
    \text{Softmax}(x_{i}) = \frac{\exp(x_i)}{\sum_{j=1}^{27} \exp(\frac{x_j}{T})}
\end{align}

The second drawback is the inability of the naive approach to correct parsing mistakes. In many cases, the model only detects at a later stage that the underlying sub-tree is incorrect, as the softmax probabilities are decreasing. One solution to determine the optimal tree would be to simply parse all possible trees and compare their calibrated softmax scores. However, this operation is computationally intensive, which is why we identify the best tree by using beam search. Specifically, we sort the parent nodes according to their softmax scores at each stage and cache a predefined number of the most promising nodes (\textit{beam width}). 

\section{Demo Plan}

\textbf{User Interface}
During the workshop we will use our online demo of \emph{CATE}. The UI consists of two parts. The left side features a text input field, where a causal sentence can be entered. In addition, the configuration of \emph{CATE} can be adjusted. It is possible to set the beam width and to select whether temperature scaling should be used for prediction. Additionally, the user can specify whether the binary tree should be built using left or right branching and which word embeddings should be utilized. After clicking on the \keys{Predict} button, the parsing result is displayed on the right side. The user can interact with the binary tree by expanding or collapsing certain segments.

\textbf{Evaluation on unseen data}
In the workshop, we will demonstrate that \emph{CATE} is suitable for practical use. For this purpose, we evaluate \emph{CATE} on unseen real word data to simulate its application in the intended context. We use a publicly available data set of acceptance criteria provided by SAP\footnote{The data set can be found at \url{https://github.com/corona-warn-app/cwa-documentation/blob/master/scoping_document.md}.}, which describe the functionality of the German ``Corona-Warn-App''. The data set consists of 32 user stories, which contain a total of 61 acceptance criteria. We found that 32 of these acceptance criteria exhibit causality~\cite{fischbach2021cira}, making them well suited for the evaluation of our approach. During the workshop, we will present both true and false predictions and discuss the performance, challenges, and possibilities of \emph{CATE} with the other participants.

\bibliographystyle{IEEEtran}
\bibliography{references}

\begin{thebibliography}{10}
\providecommand{\url}[1]{#1}
\csname url@samestyle\endcsname
\providecommand{\newblock}{\relax}
\providecommand{\bibinfo}[2]{#2}
\providecommand{\BIBentrySTDinterwordspacing}{\spaceskip=0pt\relax}
\providecommand{\BIBentryALTinterwordstretchfactor}{4}
\providecommand{\BIBentryALTinterwordspacing}{\spaceskip=\fontdimen2\font plus
\BIBentryALTinterwordstretchfactor\fontdimen3\font minus
  \fontdimen4\font\relax}
\providecommand{\BIBforeignlanguage}[2]{{%
\expandafter\ifx\csname l@#1\endcsname\relax
\typeout{** WARNING: IEEEtran.bst: No hyphenation pattern has been}%
\typeout{** loaded for the language `#1'. Using the pattern for}%
\typeout{** the default language instead.}%
\else
\language=\csname l@#1\endcsname
\fi
#2}}
\providecommand{\BIBdecl}{\relax}
\BIBdecl

\bibitem{fischbachREFSQ}
J.~Fischbach, J.~Frattini, A.~Spaans, M.~Kummeth, A.~Vogelsang, D.~Mendez, and
  M.~Unterkalmsteiner, ``Automatic detection of causality in requirement
  artifacts: the cira approach,'' in \emph{REFSQ'21}.

\bibitem{fischbachICST}
J.~Fischbach, A.~Vogelsang, D.~Spies, A.~Wehrle, M.~Junker, and
  D.~Freudenstein, ``Specmate: Automated creation of test cases from acceptance
  criteria,'' in \emph{ICST'20}.

\bibitem{fischbachRENEXT}
J.~{Fischbach}, B.~{Hauptmann}, L.~{Konwitschny}, D.~{Spies}, and
  A.~{Vogelsang}, ``Towards causality extraction from requirements,'' in
  \emph{RE'20}.

\bibitem{yang2021survey}
J.~Yang, S.~C. Han, and J.~Poon, ``A survey on extraction of causal relations
  from natural language text,'' \emph{CoRR}, vol. abs/2101.06426, 2021.

\bibitem{socher13}
R.~Socher, A.~Perelygin, J.~Wu, J.~Chuang, C.~D. Manning, A.~Ng, and C.~Potts,
  ``Recursive deep models for semantic compositionality over a sentiment
  treebank,'' in \emph{EMNLP'13}.

\bibitem{fischbachAIRE}
J.~Fischbach, T.~Springer, J.~Frattini, H.~Femmer, and A.~Vogelsang,
  ``Fine-grained causality extraction from natural language requirements using
  recursive neural tensor networks,'' in \emph{AIRE'21}.

\bibitem{socher11}
R.~Socher, C.~C.-Y. Lin, A.~Y. Ng, and C.~D. Manning, ``Parsing natural scenes
  and natural language with recursive neural networks,'' ser. ICML'11.

\bibitem{fasttext}
T.~Mikolov, E.~Grave, P.~Bojanowski, C.~Puhrsch, and A.~Joulin, ``Advances in
  pre-training distributed word representations,'' in \emph{LREC'18}.

\bibitem{pennington2014glove}
J.~Pennington, R.~Socher, and C.~D. Manning, ``Glove: Global vectors for word
  representation,'' in \emph{EMNLP'14}.

\bibitem{bert}
J.~Devlin, M.~Chang, K.~Lee, and K.~Toutanova, ``{BERT:} pre-training of deep
  bidirectional transformers for language understanding,'' \emph{CoRR}, vol.
  abs/1810.04805, 2018.

\bibitem{Guo17}
C.~Guo, G.~Pleiss, Y.~Sun, and K.~Q. Weinberger, ``On calibration of modern
  neural networks,'' \emph{CoRR}, vol. abs/1706.04599, 2017.

\bibitem{fischbach2021cira}
J.~Fischbach, J.~Frattini, and A.~Vogelsang, ``Cira: A tool for the automatic
  detection of causal relationships in requirements artifacts,'' in
  \emph{NLP4RE'21}.

\end{thebibliography}

\newpage

\onecolumn
\section{Screenshots of \emph{CATE}}

  \begin{figure}[!htb]
        \centering
         \fbox{\includegraphics[width=\linewidth]{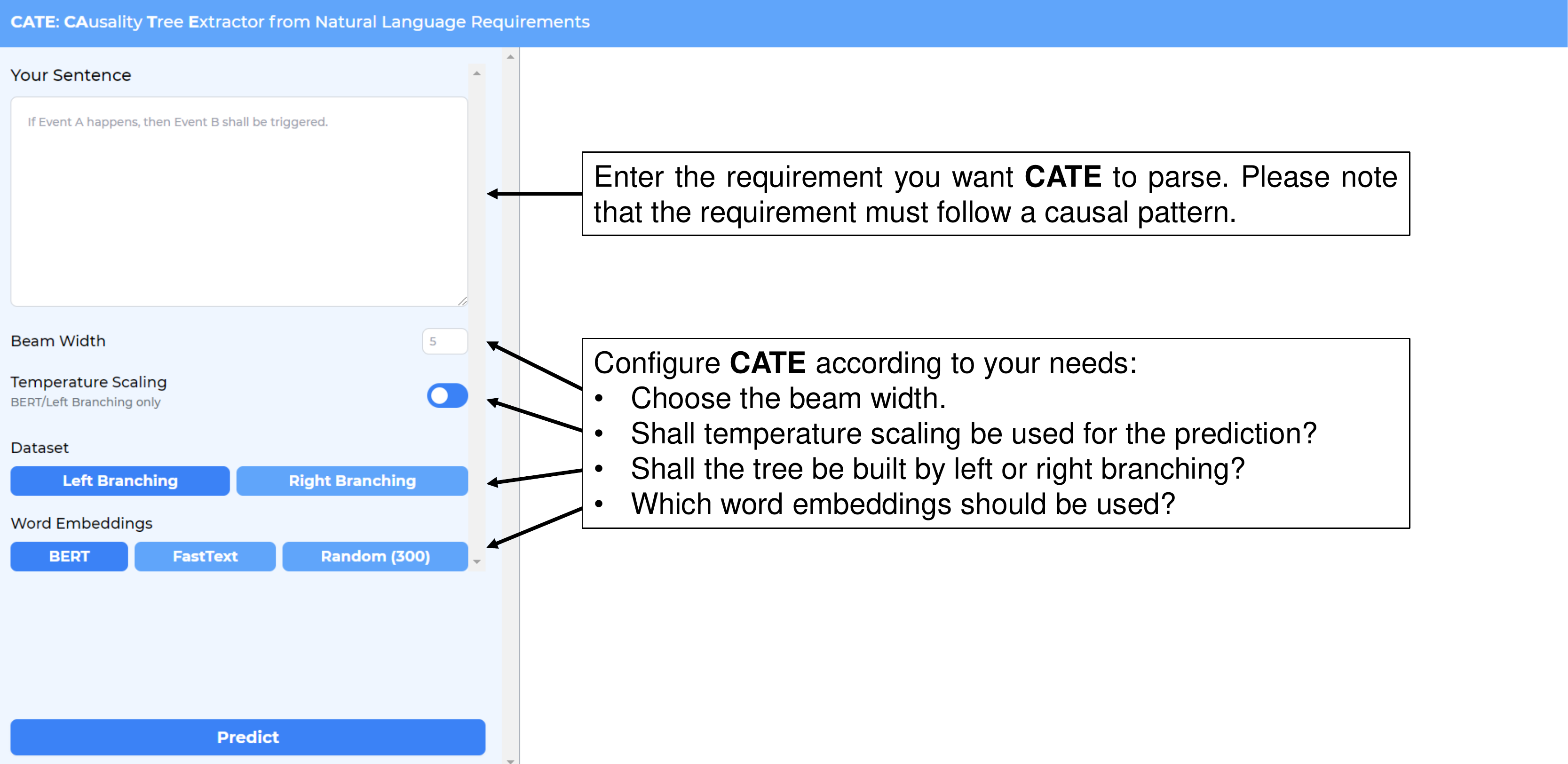}}
    \caption{Configuration options of \emph{CATE}.}\label{functionRNTN}
    \end{figure}

  \begin{figure}[!htb]
        \centering
         \fbox{\includegraphics[width=\linewidth]{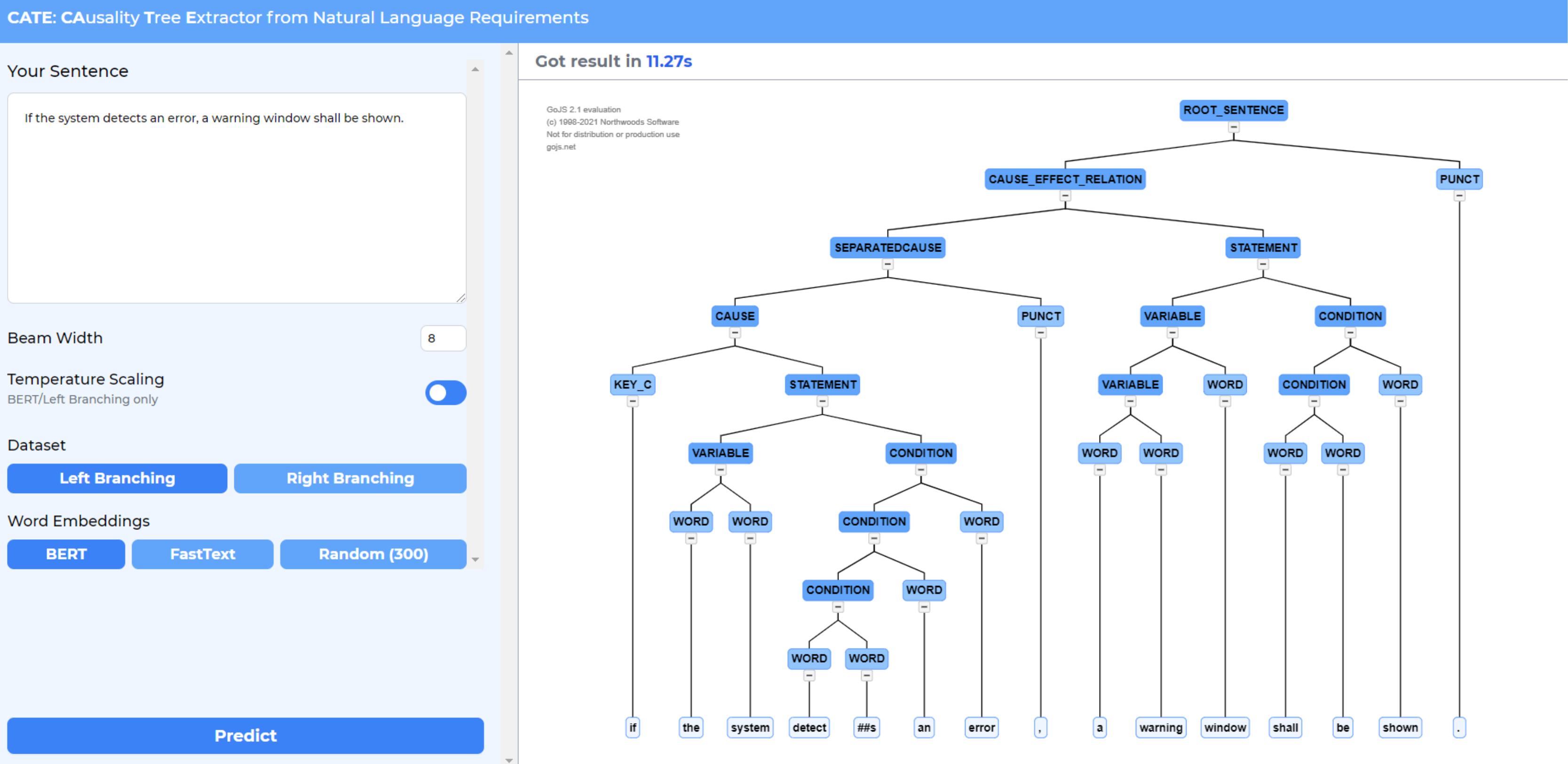}}
    \caption{Binary parse of the requirement ``\textit{If the system detects an error, a warning window shall be shown.}''}\label{functionRNTN}
    \end{figure}

\end{document}